\documentclass[10pt,twocolumn,letterpaper]{article}

\usepackage{iccv}
\usepackage{times}
\usepackage{amsmath}
\usepackage{amssymb}

\usepackage[dvipdfmx]{graphicx}
\usepackage{multirow}
\usepackage{subcaption}
\usepackage{threeparttable}
\usepackage{wasysym}

\usepackage{siunitx} 
\usepackage{booktabs} 

\usepackage{setspace}
\newcommand{\Tref}[1]{Table~\ref{#1}}

\newcommand{\Fref}[1]{Fig.~\ref{#1}}

\usepackage[pagebackref=true,breaklinks=true,letterpaper=true,colorlinks,bookmarks=false]{hyperref}

\iccvfinalcopy 

\def\iccvPaperID{1} 

\ificcvfinal\fi
\begin{document}

\title{TriDepth: Triangular Patch-based Deep Depth Prediction}

\author{Masaya Kaneko$^{1,2}$, ~~~Ken Sakurada$^{2}$, ~~~Kiyoharu Aizawa$^{1}$\\
$^{1}$The University of Tokyo, ~~~$^{2}$National Institute of Advanced Industrial Science and Technology (AIST)\\
{\tt\small $^{1}$\{kaneko, aizawa\}@hal.t.u-tokyo.ac.jp, $^{2}$k.sakurada@aist.go.jp}
}

\maketitle

\newcommand\blfootnote[1]{%
  \begingroup
  \renewcommand\thefootnote{}\footnote{#1}%
  \addtocounter{footnote}{-1}%
  \endgroup
}

\blfootnote{This work is partially supported by KAKENHI 18K18071 and JST CREST JPMJCR19F4.}

\begin{abstract}
  We propose a novel and efficient representation for single-view depth estimation using Convolutional Neural Networks (CNNs).
  Point-cloud is generally used for CNN-based 3D scene reconstruction;
  however it has some drawbacks: (1) it is redundant as a representation for planar surfaces, and (2) no spatial relationships between points are available (e.g, texture and surface).
  As a more efficient representation, we introduce a triangular-patch-cloud, 
  which represents the surface of the 3D structure using a set of triangular patches,
  and propose a CNN framework for its 3D structure estimation.
  In our framework, we create it by separating all the faces in a 2D mesh, which are determined adaptively from the input image, 
  and estimate depths and normals of all the faces.
  Using a common RGBD-dataset, we show that our representation has a better or comparable performance than the existing point-cloud-based methods, 
  although it has much less parameters.
\end{abstract}
\vspace{-5mm}

\section{Introduction}
Image-based 3D reconstruction and modeling are important problems for a variety of applications such as robotics, autonomous vehicles, and augmented reality.
The representative techniques include Structure from Motion (SfM), Multi-View Stereo (MVS), and Simultaneous Localization and Mapping (SLAM).

Recently, there have been many studies that used Convolutional Neural Networks (CNNs) for 3D reconstruction.
CNN-based single-view dense depth map prediction is a successful example~\cite{eigen2015iccv, eigen2014nips, laina2016deeper}.
In those works, the point-cloud is used for general representation since it is easy to use in CNNs, but it has some drawbacks:
(1) the parameter size is too large, and (2) the spatial relationships between the points are not described. 
On the other hand, a mesh is one representation that can solve these issues and represent 3D structures more efficiently, 
because it can simplify surfaces (e.g., room wall) and maintain the texture and surface information of the object.
However, due to the incompatibility of meshes with CNNs, conventional CNN-based approaches~\cite{kato2018renderer, wang2018pixel2mesh} cannot be used for general 3D scene reconstruction.
They are only suitable for the representation of simple 3DCG models~\cite{Chang2015ShapeNetAI}.

\begin{figure}[t]
  \centering
	\begin{minipage}{0.32\hsize}
    \includegraphics[width=1\hsize]{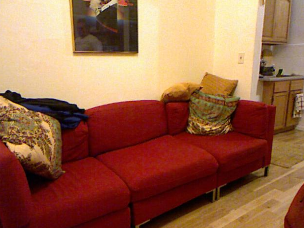}
    \subcaption{RGB Image}
  \end{minipage}
  \hfil
	\begin{minipage}{0.32\hsize}
    \includegraphics[width=1\hsize]{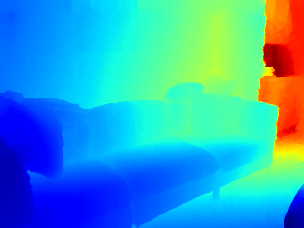}
    \subcaption{GT Depth Map}
  \end{minipage}
  \hfil
	\begin{minipage}{0.32\hsize}
    \includegraphics[width=1\hsize]{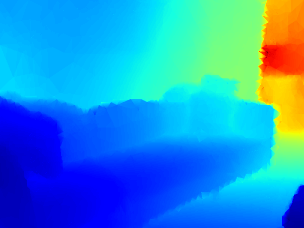}
    \subcaption{Our Depth Map}
  \end{minipage}
  \\
	\begin{minipage}{0.98\hsize}
    \includegraphics[width=0.5\hsize]{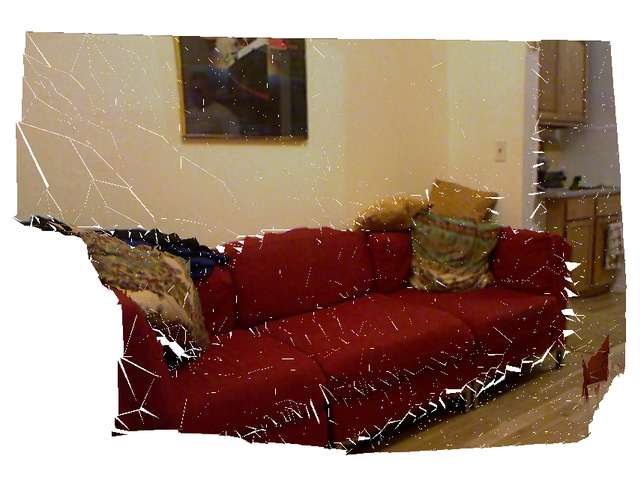}%
    \includegraphics[width=0.5\hsize]{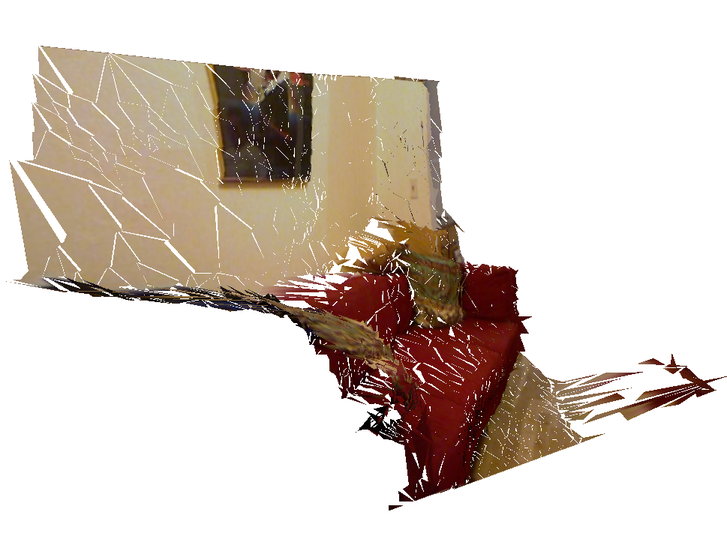}
    \subcaption{Predicted 3D structure from two views}
    \label{fig:ex2_normal}
  \end{minipage}
  \caption{We present a single-view depth prediction method using triangular-patch-cloud.
           Our representation has better performance than that of the point-cloud methods, despite having much less parameters.}
  \label{fig:result_sample1}
\end{figure}

\begin{figure*}[h]
  \begin{center}
      \includegraphics[width=0.89\hsize]{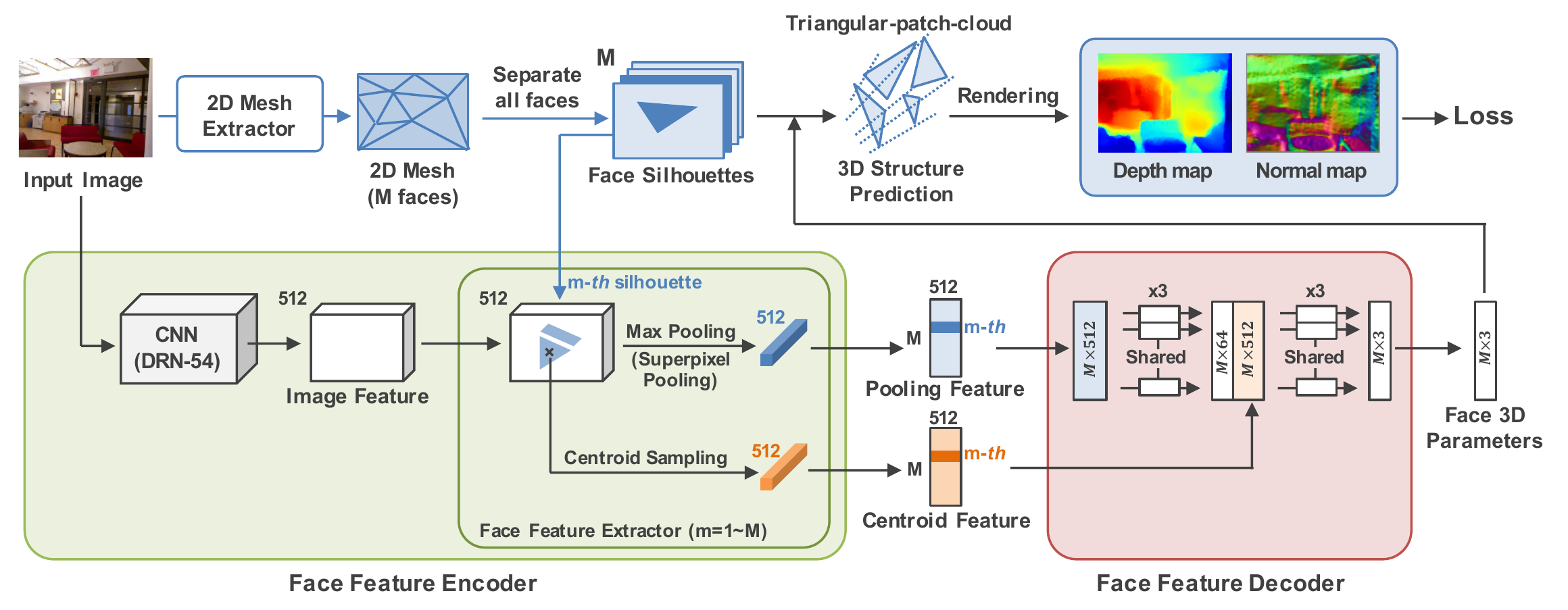}
  \end{center}
  \vspace{-5mm}
  \caption{{\bf Illustration of our framework.} After determining a 2D mesh adaptively to the input image, the CNNs estimate the 3D position of each face of the 2D mesh. 
            We train the CNNs by back-propagating the loss between the rendered results.}
  \label{fig:method:architecture}
  \vspace{-4mm}
\end{figure*}

To solve these problems, we introduce a novel intermediate representation, namely triangular-patch-cloud, between the point-cloud and mesh,
and create a novel CNN architecture for the 3D structure estimation (shown in \Fref{fig:method:architecture}).
The representation is a set of triangular patches created by separating all the faces of a 2D mesh, which is determined adaptively to the input image (in \Fref{fig:method:mesh_param}).
Since it is derived from a mesh, it has the same properties as that of the mesh representation, which means a more efficient representation than point-cloud, while still being a CNN-friendly representation. 
In our framework, we estimate the depths and normals of all the faces of the representation using CNNs, and finally obtain the 3D structure.
We evaluated the performance of our method on NYU Depth v2~\cite{Silberman2012nyudepth}.
Our method achieved better or comparable performance to the existing pixel-wise-based dense depth map estimation methods.
It should be noted that our representation has much less parameters than the existing methods.

\vspace{-2mm}

\vspace{-2mm}

\section{Proposed Method}
\label{sec:method}
\subsection{Formulation of Triangular-patch-cloud}
\begin{figure}[t]
  \begin{center}
      \includegraphics[width=0.88\hsize]{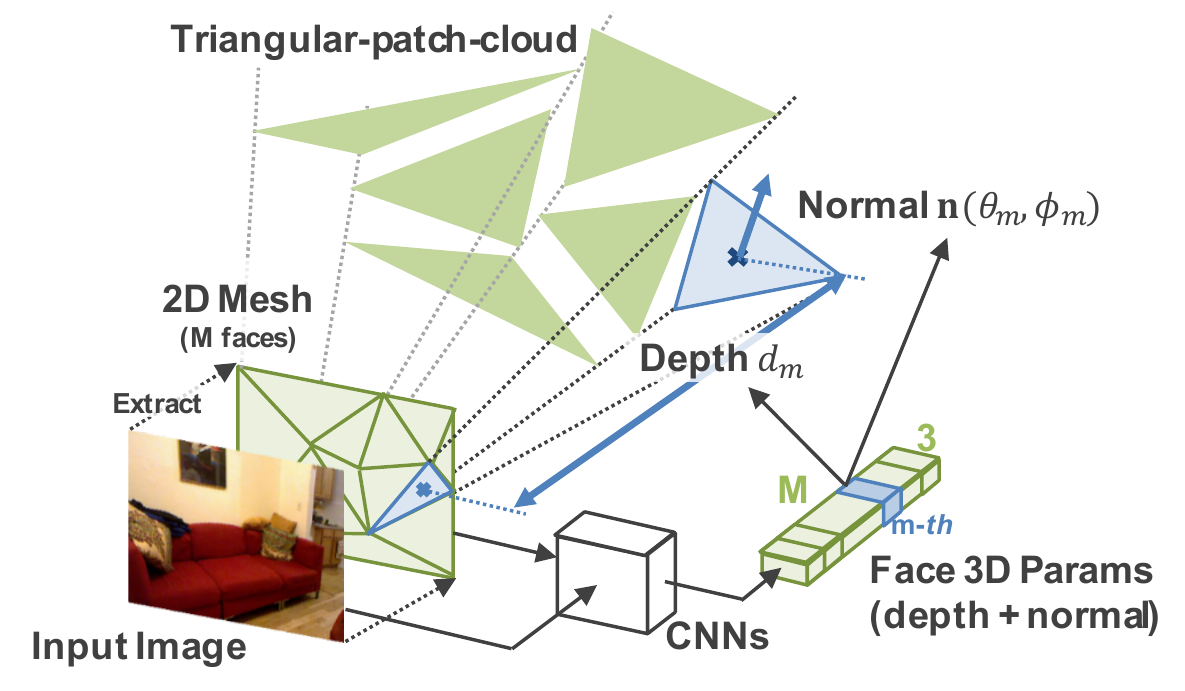}
  \end{center}
  \vspace{-5mm}
  \caption{{\bf Triangular-patch-cloud.}
           For each face, three parameters are used to determine its 3D position.}
  \label{fig:method:mesh_param}
  \vspace{-2mm}
\end{figure}

\begin{figure}[t]
  \centering
  \begin{minipage}{0.32\hsize}
    \includegraphics[width=1\hsize]{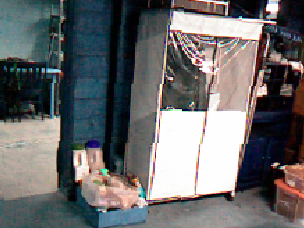}
    \subcaption{Scene Image}
  \end{minipage}
  \hfil
  \begin{minipage}{0.32\hsize}
    \includegraphics[width=1\hsize]{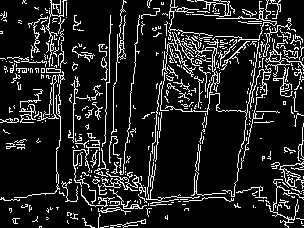}
    \subcaption{Canny Edge}
  \end{minipage}
  \hfil
  \begin{minipage}{0.32\hsize}
    \includegraphics[width=1\hsize]{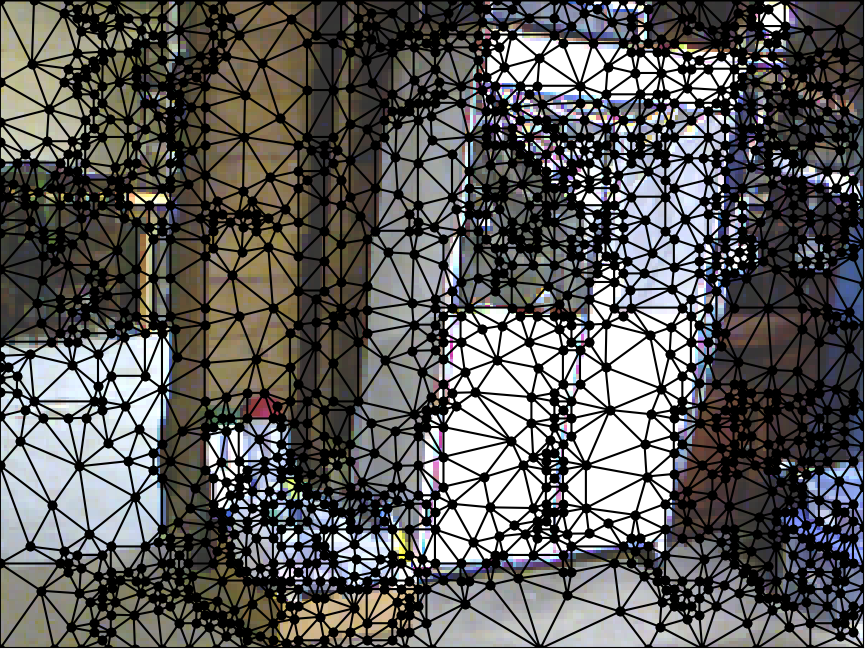}
    \subcaption{Final 2D mesh}
  \end{minipage}
  \vspace{-1mm}
  \caption{{\bf 2D Mesh Extraction.} 
           We create a 2D mesh based on (b) the Canny edge of (a) the input image.}
  \label{fig:method:mesh_extraction}
  \vspace{-6mm}
\end{figure}

\vspace{-1mm}
\label{sec:method:disconnected_mesh}
We introduce a novel representation, the triangular-patch-cloud, 
as an intermediate representation between the point-cloud and mesh, which means that it has the best of both worlds. 
It is created by ignoring the adjacency connection between the faces in a 2D mesh, 
and the faces are treated as independent triangular patches.
Each triangular patch represents a partial surface of the 3D structure.
In our framework, we first determine the base 2D mesh adaptively to the input image, and estimate the 3D positions of all the patches using CNNs, as illustrated in \Fref{fig:method:mesh_param}.
The detailed procedure of our approach is summarized as follows:
\begin{enumerate}
  \setlength{\itemsep}{5pt}      
	\setlength{\parskip}{0pt}      
	\setlength{\itemindent}{0pt}   
	\setlength{\labelsep}{5pt}     
  \item {\bf 2D mesh extraction.} We extract an appropriate 2D mesh for an input image, as shown in \Fref{fig:method:mesh_extraction}. 
  We construct partially connected vertices from the simplification of the Canny edge and obtain the final mesh by applying Constrained Delaunay Triangulation (CDT)~\cite{chew1987cdt} to these vertices.
  The 2D mesh has an adaptive number of vertices and faces to the input image.
  \item {\bf 3D structure prediction.} The faces of the obtained 2D mesh are treated independently as a triangular-patch-cloud, and the CNNs estimate the 3D positions (depths and normals) of the faces.
\end{enumerate}

\begin{figure*}[ht]
  \centering
  \begin{minipage}{0.1375\hsize}
    \includegraphics[width=1\hsize]{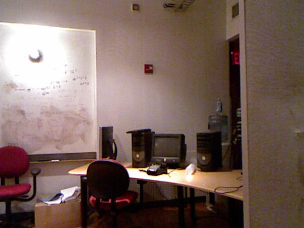} \\
    \includegraphics[width=1\hsize]{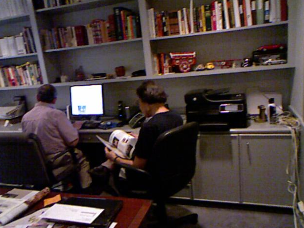} \\
    \includegraphics[width=1\hsize]{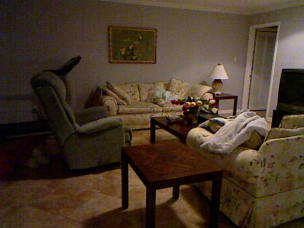}
    \subcaption{Input}
  \end{minipage}
  \hfil
  \begin{minipage}{0.1375\hsize}
    \includegraphics[width=1\hsize]{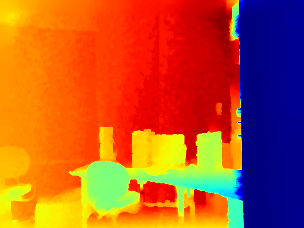} \\
    \includegraphics[width=1\hsize]{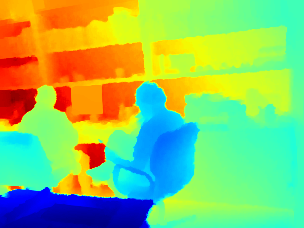} \\
    \includegraphics[width=1\hsize]{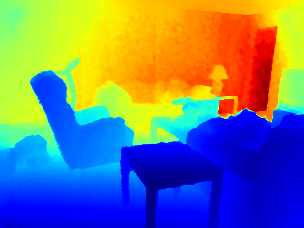}
    \subcaption{GT}
  \end{minipage}
  \hfil
  \begin{minipage}{0.1375\hsize}
    \includegraphics[width=1\hsize]{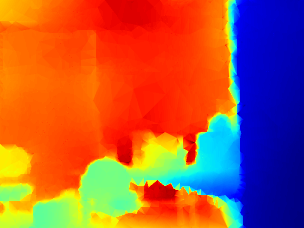} \\
    \includegraphics[width=1\hsize]{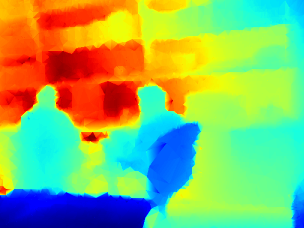} \\
    \includegraphics[width=1\hsize]{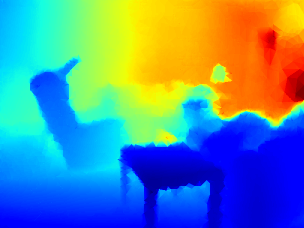}
    \subcaption{{\bf Ours}}
  \end{minipage}
  \hfil
  \begin{minipage}{0.275\hsize}
    \includegraphics[width=0.5\hsize]{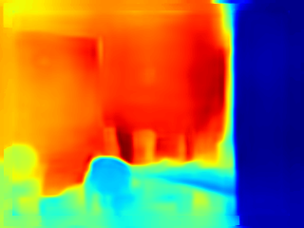}%
    \includegraphics[width=0.5\hsize]{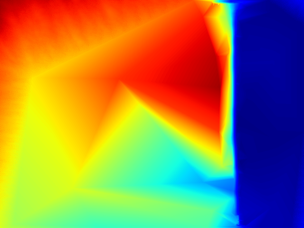}\\
    \includegraphics[width=0.5\hsize]{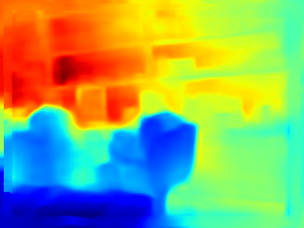}%
    \includegraphics[width=0.5\hsize]{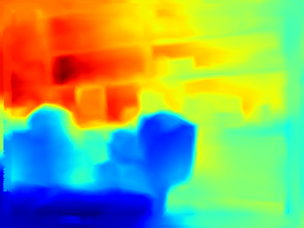}\\
    \includegraphics[width=0.5\hsize]{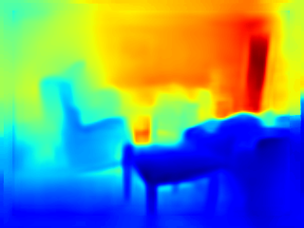}%
    \includegraphics[width=0.5\hsize]{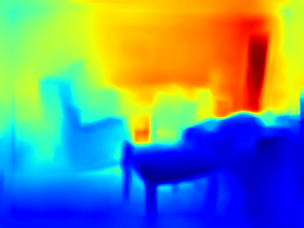}
    \subcaption{Eigen {\it et al.}~\cite{eigen2015iccv} (original, naive)}
  \end{minipage}
  \hfil
  \begin{minipage}{0.275\hsize}
    \includegraphics[width=0.5\hsize]{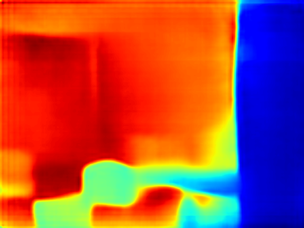}%
    \includegraphics[width=0.5\hsize]{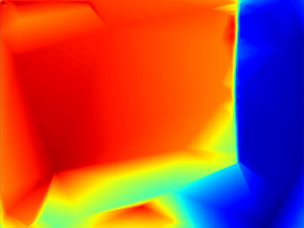}\\
    \includegraphics[width=0.5\hsize]{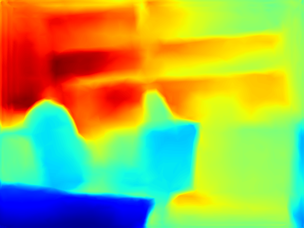}%
    \includegraphics[width=0.5\hsize]{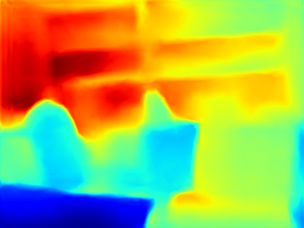}\\
    \includegraphics[width=0.5\hsize]{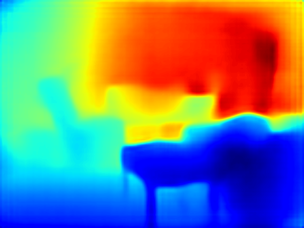}%
    \includegraphics[width=0.5\hsize]{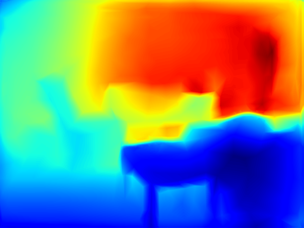}
    \subcaption{Laina {\it et al.}~\cite{laina2016deeper} (original, naive)}
  \end{minipage}
  \vspace{-2.5mm}
  \caption{{\bf Depth Map Predictions.} 
           Qualitative results showing our depth map results, the results of the pixel-wise-based methods~\cite{eigen2015iccv, laina2016deeper} (left) and their mesh-based naive methods (right).
           Each depth map is scaled for better visualization.}
  \label{fig:exp:compare_figs}
  \vspace{-5mm}
\end{figure*}

\vspace{-3mm}
\subsection{Network Architecture}
\label{sec:method:architecture}
The architecture of the proposed CNN framework for 3D structure estimation is illustrated in \Fref{fig:method:architecture}.
The framework is roughly composed of two parts:
\begin{enumerate}
  \item {\bf Face Feature Encoder.} We first extract the global feature map of the input image using DRN-54~\cite{Yu2017drn} and convert it to the face features of the prepared 2D mesh (whose number of faces $M$ is adaptive for each input image).
  For the face feature extraction, we reduce the information from one feature vector per pixel to one feature vector per face region (silhouette). 
  We adopt two methods: (1) Superpixel Pooling~\cite{kwak2017spp}, which extract the max values in the region, and (2) Face Centroid Sampling, which extracts the value on the centroid position of the region.
  \item {\bf Face Feature Decoder.} By using the features --- face pooling feature and face centroid feature ---, we finally estimate the parameters representing the 3D position of each face. 
  A patch of the triangular-patch-cloud has similar properties as that of a point in a point-cloud because both of them are unordered and interact with each other. 
  Therefore, we created a CNN composed of a shared multi-layer perceptron (MLP) network, which is similar to PointNet~\cite{Qi2017pointnet}.
\end{enumerate}
\vspace{-5mm}

\subsection{Loss Function}
\label{sec:method:loss_function}
\vspace{-1.5mm}
Here, we explain the method to optimize the proposed CNN framework.
We use general RGBD datasets for the training of our framework and define the loss ($L_{depth}$) between the depth map $D$ rendered by Neural Mesh Renderer~\cite{kato2018renderer} and its ground truth (GT) depth map $D^*$.
In addition, we include the normal loss ($L_{normal}$) between the corresponding normal map $N$ and the GT normal map $N^*$ calculated by~\cite{Yang2018normal} in our loss function, as follows.
\vspace{-1mm}
\begin{align}
  \displaystyle L_{sum} &= L_{depth} + \lambda_{n} L_{normal} \notag \\
                  &= \frac{1}{n}\sum_{i}(D_{i}-D_{i}^*) + \lambda_{n} (-\frac{1}{n}\sum_{i}(N_{i}\cdot N_{i}^*)) 
  \label{eq:loss}
\end{align}
where $i$ is the valid pixel id, 
$n$ is the total number of valid depth pixels, 
and $\lambda_{n}$ is a balancing factor (we use $\lambda_{n}=0.5$ as the best value).

\vspace{-2mm}

\section{Experimental Results}
\vspace{-1.5mm}
\label{sec:exp}
For the evaluation of our proposed approach, we trained our framework using NYU depth v2~\cite{Silberman2012nyudepth}, which is one of the largest RGBD datasets for indoor scene reconstruction.
This dataset is composed of pairs of an RGB image and the depth image of 464 scenes captured by Microsoft Kinect.
We followed the official splitting, i.e., 249 scenes for training and 215 scenes for testing.
For this evaluation, we used approximately 48K pairs, which were sampled spatially uniformly from the scenes in the raw training dataset for training, 
and used 654 labelled images for evaluating the final performance.
The input image to the network was resized to $228 \times 304$ following previous works~\cite{eigen2015iccv,laina2016deeper, Ma2017SparseToDense}.
We augmented them with some random transformations (small rotations, scaling, color jitter, color normalization, and flips with 0.5 chance)~\cite{Ma2017SparseToDense}.

We trained our method for approximately 50 epochs with a batch size of 4 on a single NVIDIA Tesla P40 with 24GB of GPU memory.
We used $1\%$ of the training dataset separately as a validation dataset for the hyperparameter search.
The final score of the test dataset was evaluated using the trained model with the highest validation score.
As the quantitative evaluation metrics, we used the general error metrics (RMSE, REL, $\log_{10}$, $\delta_1 \sim \delta_3$)~\cite{eigen2015iccv, laina2016deeper} for performance comparison.

\begin{table}[t]
	\centering
    \scriptsize
    {\tabcolsep=1.3mm
    \caption{{\bf Comparison with point-cloud-based methods.}
             It is advantageous to have low error metrics (REL, RMSE, $\log_{10}$) and high accuracy metrics ($\delta_1 \sim \delta_3$).}
    \vspace{-2mm}
    \begin{tabular}{|l||c|c|c|c|c|c|c|} \hline
      Method                                           & rel         & rms         & $\log_{10}$ & $\delta_1$  & $\delta_2$  & $\delta_3$  & \#param.  \\ \hline \hline
      Eigen and Fergus~\cite{eigen2015iccv}            & 0.158       & 0.641       & -           & 0.769       & 0.950       & {\bf 0.988} & 921K      \\
      Laina~\cite{laina2016deeper}                     & {\bf 0.127} & 0.573       & {\bf 0.055} & {\bf 0.811} & 0.953       & {\bf 0.988} & 921K      \\ \hline \hline
      {\bf Ours}                                       & 0.146       & {\bf 0.530} & 0.062       & 0.803       & {\bf 0.954} & {\bf 0.988} & {\bf 32K} \\ \hline
    \end{tabular}
	  \label{table:exp:compare_pixel}
    }
\vspace{-2mm}
\end{table}

\begin{table}[t]
	\centering
    \scriptsize
    {\tabcolsep=1.3mm
    \caption{{\bf Comparison with naive mesh-based methods.}
              For each method, we use the results provided by the authors.}
    \vspace{-2mm}
    \begin{tabular}{|l||c|c|c|c|c|c|c|} \hline
      Method                    & rel         & rms         & $\log_{10}$ & $\delta_1$  & $\delta_2$  & $\delta_3$ \\ \hline \hline
      ver. Eigen and Fergus~\cite{eigen2015iccv} & 0.163       & 0.559       & 0.069       & 0.762       & 0.948       & 0.987      \\
      ver. Laina~\cite{laina2016deeper}          & 0.154       & 0.535       & 0.064       & 0.793       & 0.949       & 0.987      \\ \hline \hline
      {\bf Ours}                            & {\bf 0.146} & {\bf 0.530} & {\bf 0.062} & {\bf 0.803} & {\bf 0.954} & {\bf 0.988}\\ \hline
    \end{tabular}
	  \label{table:exp:naive_method}
    }
\vspace{-5mm}
\end{table}

\vspace{-1mm}
\subsection{Comparison with Point-cloud-based methods.}
\begin{figure}[h]
  \centering
  \begin{minipage}{0.98\hsize}
    \includegraphics[width=0.3333\hsize]{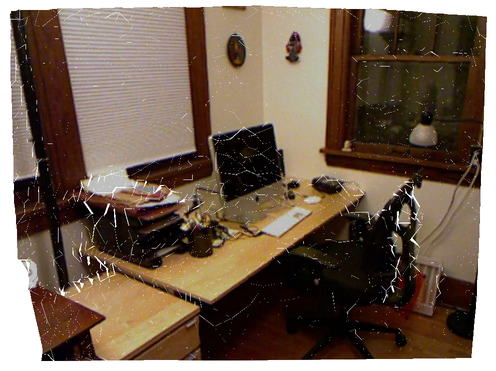}%
    \includegraphics[width=0.3333\hsize]{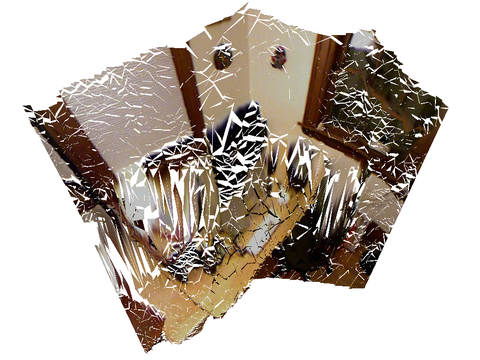}%
    \includegraphics[width=0.3333\hsize]{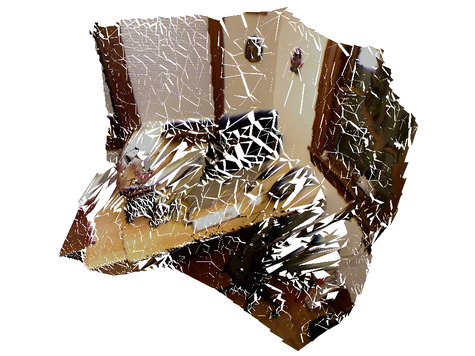}
    \includegraphics[width=0.3333\hsize]{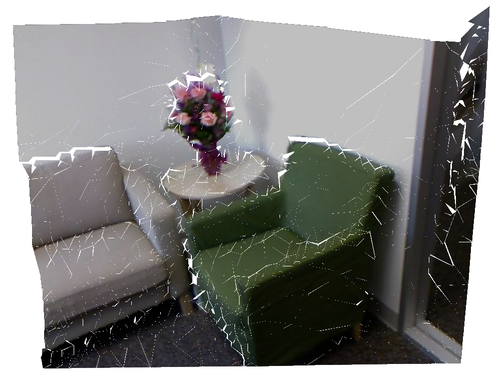}%
    \includegraphics[width=0.3333\hsize]{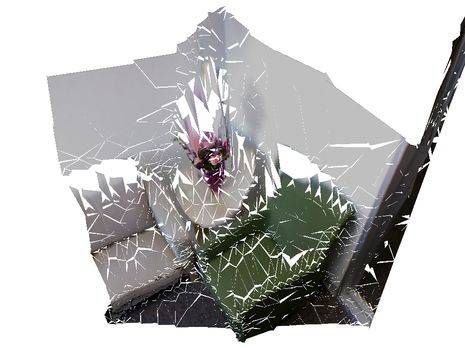}%
    \includegraphics[width=0.3333\hsize]{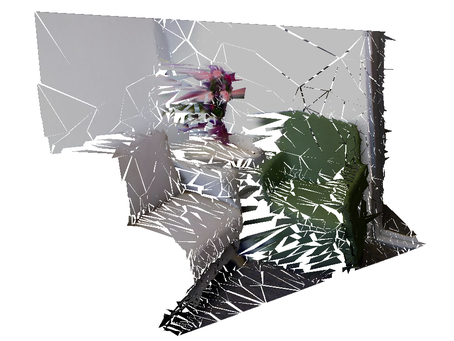}
    \includegraphics[width=0.3333\hsize]{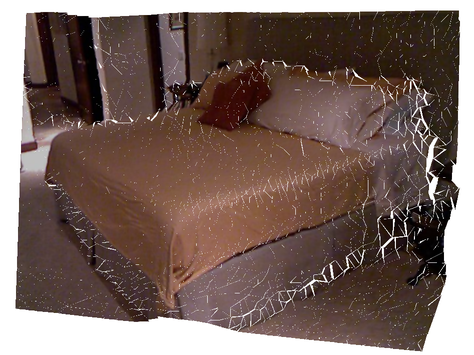}%
    \includegraphics[width=0.3333\hsize]{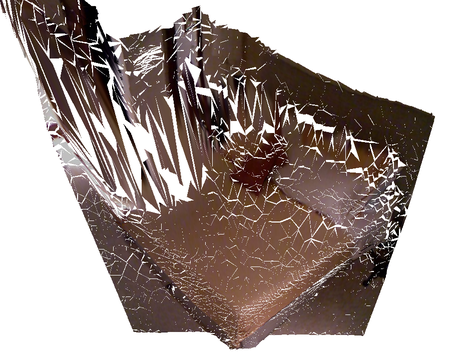}%
    \includegraphics[width=0.3333\hsize]{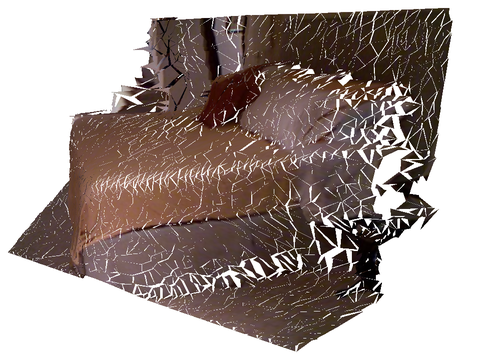}
  \end{minipage}
  \vspace{-1mm}
  \caption{{\bf 3D Predictions of Triangular-patch-cloud.}
            Visualization results of our predictions from multi views.}
  \label{fig:exp:3d_mesh}
  \vspace{-5mm}
\end{figure}

\vspace{-1mm}
We compare our method with the baseline methods of dense depth map prediction (point-cloud-based methods).
To perform an evaluation similar to the point-cloud-based methods, we use the depth map rendered from the estimated 3D mesh (see \Fref{fig:exp:3d_mesh}).
The results are provided in \Tref{table:exp:compare_pixel}.
The performance of our method was better or comparable to that of the existing methods, despite having much less parameters.
The parameter size implies the size of the 3D points registered in the 3D map in SfM and visual SLAM.

\vspace{-1mm}
\subsection{Comparison with Naive mesh-based methods.}
\vspace{-1mm}
Next, we compare the performance of our method with those of the naive mesh-based methods, since there are no CNN-based 3D scene mesh reconstruction methods.
Here, a naive method involves the following procedure:
(1) A dense mesh is constructed by connecting the adjacent vertices of the dense depth map obtained by a point-cloud-based methods~\cite{eigen2015iccv, laina2016deeper}.
(2) The dense mesh is simplified until it has the same number of faces as that in our method.
This process gives the results rendered by a mesh with a similar parameter size.

The results are shown in \Tref{table:exp:naive_method}.
Our method achieved better accuracy than both methods for all the evaluation metrics.
Furthermore, according to the qualitative results displayed in \Fref{fig:exp:compare_figs}, 
the naive methods could only estimate rough and poor depth maps (see the top line),
but our method could estimate the depth maps that reflected the object boundaries clearly.
This results demonstrate that the depth map prediction based on our representation is effective for complex 3D scenes.

\vspace{-2.5mm}

\section{Conclusions}
\vspace{-1.5mm}
We presented a novel approach for CNN-based 3D scene reconstruction of complex indoor scenes, using intermediate representation between the mesh and point-cloud.
Our representation is CNN-friendly and more efficient than point-cloud.
We showed that our framework could predict a visually clean 3D structure, and the results indicated equal or better performance than that of the existing methods even though our representation had much less parameters.

As a future work, we plan to partially connect the faces that are completely separated in this work.
The triangular-patch-cloud is created by separating all the faces of a 2D mesh to represent complex shapes (especially occlusions); 
however, there are many faces that should be connected.
By connecting the faces partially, we will create a mesh, a so-called partially-disconnected-mesh, whose appearance would surely become clearer while the number of parameters are reduced further. 

\begin{figure}[t]
  \centering
  \begin{minipage}{0.31\hsize}
    \includegraphics[width=1\hsize]{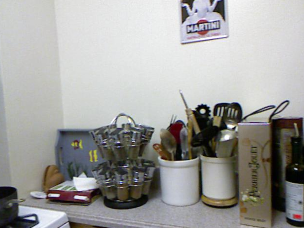} \\
    \includegraphics[width=1\hsize]{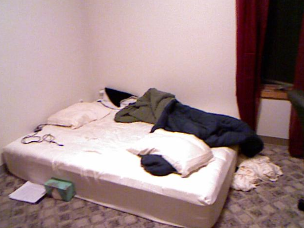} \\
    \includegraphics[width=1\hsize]{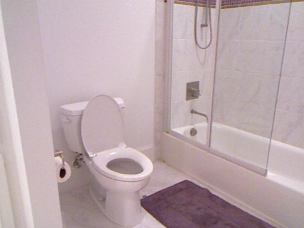}
    \subcaption{Input}
  \end{minipage}
  \hfil
  \begin{minipage}{0.31\hsize}
    \includegraphics[width=1\hsize]{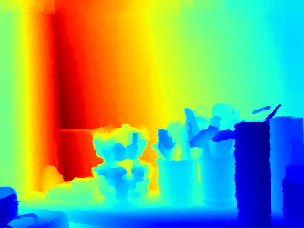} \\
    \includegraphics[width=1\hsize]{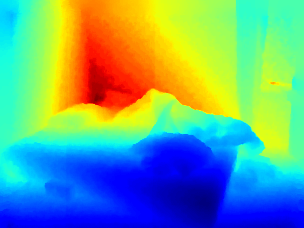} \\
    \includegraphics[width=1\hsize]{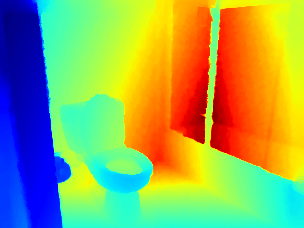}
    \subcaption{GT}
  \end{minipage}
  \hfil
  \begin{minipage}{0.31\hsize}
    \includegraphics[width=1\hsize]{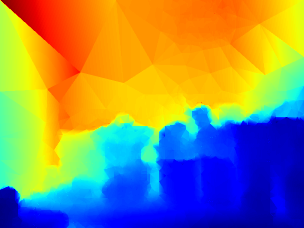} \\
    \includegraphics[width=1\hsize]{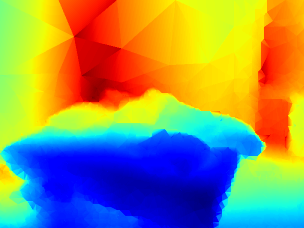} \\
    \includegraphics[width=1\hsize]{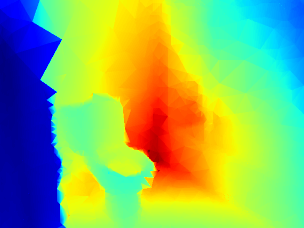}
    \subcaption{Predictions}
  \end{minipage}
  \vspace{-1mm}
  \caption{{\bf Failure Cases.}
           The current approach cannot estimate the 3D structure correctly in the scenes where the geometric edges are hard to detect from RGB information.}
  \label{fig:conclusion:limit}
  \vspace{-4mm}
\end{figure}

In addition, we will solve the limitation of our method by improving the 2D mesh extraction process.
Currently the process is created based on the Canny edge, 
which means that it does not work well in the scenes where it is difficult to recognize the geometric boundary from the RGB information (shown in \Fref{fig:conclusion:limit}).
We plan to make 2D mesh extraction trainable from a dataset in order to improve their robustness.

{\small
\bibliographystyle{ieee}
\bibliography{reference}
}

\end{document}